\documentclass[sigconf]{acmart}
\makeatletter
\def\@ACM@checkaffil{
    \if@ACM@instpresent\else
    \ClassWarningNoLine{\@classname}{No institution present for an affiliation}%
    \fi
    \if@ACM@citypresent\else
    \ClassWarningNoLine{\@classname}{No city present for an affiliation}%
    \fi
    \if@ACM@countrypresent\else
        \ClassWarningNoLine{\@classname}{No country present for an affiliation}%
    \fi
}
\makeatother

\AtBeginDocument{%
  \providecommand\BibTeX{{%
    \normalfont B\kern-0.5em{\scshape i\kern-0.25em b}\kern-0.8em\TeX}}}

\copyrightyear{2023} 
\acmYear{2023} 
\setcopyright{acmlicensed}
\acmConference[KDD '23] {Proceedings of the 29th ACM SIGKDD Conference on Knowledge Discovery and Data Mining}{August 6--10, 2023}{Long Beach, CA, USA.}
\acmBooktitle{Proceedings of the 29th ACM SIGKDD Conference on Knowledge Discovery and Data Mining (KDD '23), August 6--10, 2023, Long Beach, CA, USA}
\acmPrice{15.00}
\acmISBN{979-8-4007-0103-0/23/08} 
\acmDOI{10.1145/3580305.3599916}
\usepackage{todonotes} 

\newcommand{\tup}[1]{\left( #1 \right)}

\usepackage{amsmath}
\DeclareMathOperator{\sign}{sign}

\usepackage{algorithm}
\usepackage{algpseudocode}
\usepackage{tabularx}
\usepackage{multirow}
\usepackage{hhline}
\usepackage{float}
\usepackage{caption}
\usepackage{subcaption}
\usepackage{pgfplots}

\begin{document}

\title{Towards Fairness in Personalized Ads Using Impression Variance Aware Reinforcement Learning}



\author{Aditya Srinivas Timmaraju}
\authornote{Equal contribution from engineers}
\affiliation{%
  \institution{Meta}
}
\email{adityatimmaraju@meta.com}

\author{Mehdi Mashayekhi}
\authornotemark[1]
\authornote{Corresponding author}
\affiliation{%
  \institution{Meta}
  \streetaddress{P.O. Box 1212}
  \postcode{43017-6221}
}
\email{mmashayekhi@meta.com}

\author{Mingliang Chen}
\authornotemark[1]
\affiliation{%
  \institution{Meta}
}
\email{chenml1990@meta.com}

\author{Qi Zeng}
\authornotemark[1]
\affiliation{%
  \institution{Meta}
}
\email{qizeng@meta.com}

\author{Quintin Fettes}
\authornotemark[1]
\affiliation{%
  \institution{Meta}
}
\email{qfettes@meta.com}

\author{Wesley Cheung}
\authornotemark[1]
\affiliation{%
  \institution{Meta}
}
\email{wcheung8@meta.com}

\author{Yihan Xiao}
\authornotemark[1]
\affiliation{%
  \institution{Meta}
}
\email{yihanxiao@meta.com}

\author{Manojkumar Rangasamy Kannadasan}
\affiliation{%
  \institution{Meta}
}
\email{mkannadasan@meta.com}

\author{Pushkar Tripathi}
\affiliation{%
  \institution{Meta}
}
\email{pushkart@meta.com}

\author{Sean Gahagan}
\affiliation{%
  \institution{Meta}
}
\email{smg@meta.com}

\author{Miranda Bogen}
\affiliation{%
  \institution{Meta}
}
\email{mbogen@meta.com}

\author{Rob Roudani}
\affiliation{%
  \institution{Meta}
}
\email{rroudani@meta.com}


\renewcommand{\shortauthors}{Mashayekhi et al.}

\begin{abstract}
Variances in ad impression outcomes across demographic groups are increasingly considered to be potentially indicative of algorithmic bias in personalized ads systems. 
While there are many definitions of fairness that could be applicable in the context of personalized systems, we present a framework which we call the Variance Reduction System (VRS) for achieving more equitable outcomes in Meta’s ads systems.
VRS seeks to achieve a distribution of impressions with respect to selected protected class (PC) attributes that more closely aligns the demographics of an ad's eligible audience (a function of advertiser targeting criteria) with the audience who sees that ad, in a privacy-preserving manner.
We first define metrics to quantify fairness gaps in terms of ad impression variances with respect to PC attributes including gender and estimated race. We then present the VRS for re-ranking ads in an impression variance-aware manner.  
We evaluate VRS via extensive simulations over different parameter choices and study the effect of the VRS on the chosen fairness metric. We finally present online A/B testing results from applying VRS to Meta’s ads systems, concluding with a discussion of future work. 
We have deployed the VRS to all users in the US for housing ads, resulting in significant improvement in our fairness metric.
VRS is the first large-scale deployed framework for pursuing fairness for multiple PC attributes in online advertising. 
\end{abstract}





\received{2 February 2023}

\maketitle

\section{Introduction}
Online personalized advertising is often very effective at providing relevant content to people, based on advertisers’ preferences and the power of ranking algorithms. This has led to its pervasive adoption. More importantly, these ads systems have been used to promote important life opportunities such as housing, employment, and credit. As a result, fairness in personalized ads has emerged as an area of significant focus for policymakers, regulators, civil rights groups, industry and other stakeholders \citep{24cf, fb_civil, ali2019discrimination}, sparking calls to mitigate potential algorithmic bias with respect to protected class (PC) attribute(s) (e.g., gender, age and race).

Across both the research community and industry, approaches to fairness and inclusivity in the use of AI are still evolving, particularly in the realm of personalized, auction-based advertising systems. Even experts have struggled to articulate clear expectations for what bias or fairness ought to mean in particular contexts, especially when faced with contradictory definitions or expectations. This work is an attempt to make progress in addressing important concerns about  potential bias in Meta's ads systems especially when it comes to housing, employment, and credit.

In this paper, we present the Variance Reduction System (VRS), a framework for achieving equitable outcomes in Meta's ads systems by more closely aligning the demographics of an ad’s eligible audience (defined by advertiser targeting) and the audience that sees the ad for defined ad verticals. 
Given fairness requirements in this context have been expressed in terms of desired impression outcomes for housing ads over PC attributes (i.e., gender and race), we propose a solution to adjust the ranking of housing ads for certain users without directly using those PC attributes at an individual level. 
Our key contributions include:
\begin{itemize}
 \item Designing fairness-aware ranking systems towards mitigating algorithmic bias in systems, for cases where bias is defined as the presence of inequitable outcomes (or "variances") in impression delivery across PC attributes.
 \item Extensive evaluation of the proposed system via simulations over a wide range of PC attributes with different cardinalities.
 \item Online A/B test results of applying our system for pursuing equitable outcomes in Meta's ads stack. Our solution results in a significant improvement in the defined fairness metric.
 \end{itemize}
 
 The rest of the paper is organized as follows. In Section ~\ref{sec:Bias and Fairness Consideration} we define our choice of fairness metric and how to measure it in privacy-preserving manner. In Section ~\ref{sec:Modeling} we describe our impression variance aware modeling approach. In Section ~\ref{sec:offline_simulation} we describe our simulation environment.  In Section ~\ref{sec:Results} we present our offline and online A/B testing results.
 In Section ~\ref{sec:Related Work} we provide an overview of related work. Finally in Section ~\ref{sec:Discussion} we provide a summary of our contributions and an outlook of future work. To avoid conflation of terms, any references to "variance" will mean impression/delivery variance (as defined in Equation ~\ref{eqn_skew}) and we will explicitly use "statistical variance" when referring to the notion of variance from probability and statistics.

\section{Fairness and Privacy Considerations}
\label{sec:Bias and Fairness Consideration}

\subsection{Mapping Fairness to Desired Distribution}

Our measurement and mitigation methodology assume that in the ideal setting, the impression outcomes of housing ads over PC attributes (e.g., gender and race), should follow a desired distribution which we call the \emph{baseline} or the \emph{eligible} distribution. Researchers have proposed different dimensions of fairness, bias, and discrimination both broadly in the context of machine learning as well as more specifically in the context of online advertising. In this section, we discuss how the VRS relates to several notions of fairness.

\subsubsection{Equal Opportunity}
 Equal opportunity means that individuals who qualify for a desirable outcome should have an equal chance of being correctly classified for this outcome \cite{hardt2016equality}. In other words the desired outcome should not depend on a protected attribute such as race or gender. Formally, a predictor $\hat{Y}$ is said to satisfy equal opportunity with respect to a PC attribute $A$ and true outcome $Y$, if the predictor and the PC attribute are independent conditioned on the true outcome being 1. That is:
 \begin{equation} \label{equal_oppor}
	 p(\hat{Y} =1 \mid A=1, Y =1) =  p(\hat{Y} =1 \mid A=0, Y =1)
\end{equation}

\subsubsection{Demographic Parity}
Demographic parity requires that the prediction must be uncorrelated with sensitive attributes \cite{DemographicParity}. Formally, a predictor $\hat{Y}$ is said to satisfy demographic parity if Equation ~\ref{Demographic_Parity} holds.
 \begin{equation} \label{Demographic_Parity}
    \begin{split}
	    p(\hat{Y} =1 \mid A=1) =  p(\hat{Y} =1 \mid A=0), and \\
	    p(\hat{Y} =0 \mid A=1) =  p(\hat{Y} =0 \mid A=0)
    \end{split}
\end{equation}
The key difference between the demographic parity and equal opportunity notions of fairness is that the former does not consider qualification of the candidates and only enforces non-discrimination at a global level \cite{Sahin2019}.
\subsubsection{Equitable Outcomes}
\label{sec:equitable_outcome}
Both the equal opportunity and the demographic parity notions are formulated in the context of classification tasks. 
In the context of the complex processes such as auction-based advertising system, we aim to achieve Equitable Outcomes.
An equitable outcome in this case requires that the fraction of impressions belonging to a given value of the PC attribute does not  significantly vary between that of the eligible users and that of the delivered users (see mathematical definitions in Section ~\ref{sssec:elligible}).

The notion of Equitable Outcomes can be roughly mapped to either Equal Opportunity or Demographic Parity notions above, depending on the view of the advertiser's targeting that generates the eligible audience of an ad (see Section ~\ref{sssec:elligible}).
For example, by viewing the targeting as a qualification process for the ad delivery task, the true outcome being positive ($Y$ = 1) corresponds to a user matching the targeting criteria (or “qualified” for receiving the impression), and the prediction being positive ($\hat{Y}$ = 1) corresponds to a user actually receiving the ad impression.
Then, equitable outcome meets the requirement of equal opportunity because given an individual’s inclusion in an eligible audience, that individual has an equal chance of being delivered an ad impression regardless of their PC attribute values.
Alternatively, if one views the targeting as a rough specification of users for receiving the ad impression as opposed to a qualification process (e.g., a targeted user may still not be interested in the housing ad), equitable outcome meets the requirement of demographic parity since qualification is not taken into account.

\subsection{Variance between Baseline (Eligible) and Delivery distributions (i.e., Equitable Outcome)}
\label{sssec:elligible}
The baseline can be determined in many ways including, but not limited to, adhering to the corresponding distribution over a baseline population, a legal mandate, or a voluntary commitment \cite{EUdiversity}. 

In particular, we choose a baseline based on the eligible audience generated by an advertiser's targeting criteria for a given ad, which are already constrained by policies and standards for Housing, Employment and Credit verticals \cite{audience_hec}. When an advertiser sets up an advertisement campaign, they need to select a set of criteria (aka "targeting rules") which specify the audience that the ad is intended to reach. This ensures that ads can be delivered to the people that the advertiser can actually provide their product and service to (for example, a geographic region), and they believe will most likely be interested in their offering.

Since advertiser targeting is already constrained by policy to help prevent discriminatory targeting behavior (for example, disallowing the use of gender criteria for employment ads) and the resulting eligible audience is independent of ranking algorithms, a given ad's targeting parameters represent a reasonable baseline metric. On the other hand, since targeting rules still reflect advertiser preferences, ads will still be reasonably relevant after we impose the constraint that ad delivery outcomes more closely resemble the baseline distribution.

After an advertiser selects their target audience, the ads ranking system delivers ads to a subset of the eligible users, for example, when they spend time on Facebook or Instagram where opportunities to show an ad are available. We define \emph{delivery distribution} as the demographic distribution of ad impressions over users who receive those impressions. A sizeable \emph{variance} between the baseline and delivery distribution might indicate unintentional unfairness within the ads ranking system, so our goal is to alleviate that \emph{variance} as defined formally below.
\begin{definition}\label{variance_def}
  Variance Definition.
\end{definition}
\begin{equation} \label{eqn_delivery}
	\mbox{\emph{Delivery ratio}}_{pc, ad} = \frac{Imps_{pc, ad}}{Imps_{ad}}
\end{equation}
\begin{equation} \label{eqn_eligible}
	\mbox{\emph{Eligible ratio}}_{pc, ad} = \frac{\sum_{user \in ad\char`_target\char`_audience \bigcap pc} Imps_{user}}{\sum_{user \in ad\char`_target\char`_audience} Imps_{user}}
\end{equation}
\begin{equation} \label{eqn_skew}
	\emph{Variance}_{pc, ad} = SD(\mbox{\emph{Delivery ratio}}_{pc, ad} - \mbox{\emph{Eligible ratio}}_{pc, ad})
\end{equation}
where \emph{Imps} refers to number of impressions, and SD refers to the \emph{Shuffle Distance}.
Shuffle Distance is the minimum fraction that needs to be moved (or shuffled , hence the name) from an actual distribution $p$ (in our use case this translate to the actual delivery ratio), to match a reference distribution $\pi$ (in our use case this translate to the eligible ratio). 
Mathematically, Shuffle Distance can be calculated as half of $L_{1}$ distance between distribution $p$ and $\pi$:
\begin{equation} \label{shuffle}
    \mbox{\emph{Shuffle Distance}} = \frac{\| p - \pi \| }{2}
\end{equation}
Intuitively, a shuffle distance of 10\% means that if 10\% of impressions were moved from males to females , we would achieve a variance of 0.

\subsection{Differential Privacy}
\label{sec:diff_privacy}
In an effort to center user privacy, we built the VRS to accommodate circumstances where the model might not have access to individual user PC subgroup information at both training and inference time. On the other hand, accessing user's subgroup information is necessary when measuring variance according to Definition ~\ref{variance_def}. To address this, variance is aggregated and measured at ad level. Then, we adopt differential privacy (DP) \citep{Hilton2012DifferentialP} in order to address various common issues such as privacy attacks as discussed in~\citep{dwork2017exposed}.

By using randomness in a controlled way, such as through the $\epsilon$-differential privacy setup \citep{10.1007/11681878_14, dwork2014algorithmic}, differential privacy provides a mathematically rigid framework to quantify the desired level of protection against potential information leakage.
Specifically, we apply the \emph{additive noise mechanism} as introduced in \citep{10.1007/11681878_14}. In short, these methods add a noise drawn from Laplace or Gaussian distribution on top of the measurement on a collection of data. Formally, given a measurement function $f(x)$ over a collection of dataset $\mathcal{D}$, this mechanism will add a noise taken from Gaussian, for instance, defined by
\begin{equation} \label{dp_gaussian_noise}
	\mathcal{N}(\mu=0, \sigma^2=\dfrac{2\ln{(1.25/\delta)}\cdot(\Delta f)^2}{\epsilon^2})
\end{equation}
in order to provide $(\epsilon, \delta)$-differential privacy for $\epsilon\in(0, 1)$ and $\delta\in(0, 1)$. Here $\Delta f=max|f(x) - f(y)|$, is called \emph{sensitivity}, quantifying maximum difference over all possible pairs of $x$ and $y$ in $\mathcal{D}$ differing in at most one element. Readers are advised to refer to \cite{dwork2014algorithmic} for rigorous mathematical proof.

Additionally, a few customizations are applied to help accommodate some VRS-specific behaviors:

\begin{itemize}
    \item In VRS, we apply additive noise mechanism to the impression counts breakdown by users' subgroups, i.e. $f(x)$, as they are the ingredients in variance calculation. As a result, \emph{sensitivity} is $1$.
    \item Due to the nature of streaming data, we need to chunk ad impression data into batches and apply the \emph{additive noise mechanism} to each batch independently. These noisy counts are successively added up for the final counts. This is necessary because the system would otherwise be performing repeated measurement on a part of data, which might reduce the efficacy of privacy measures as mentioned above. Compared to naive one-time measurement on full data, this will increase the statistical variance on final count by factor of $N$, which is the number of batches. As a result, the relative uncertainty will vanish following $1/\sqrt{N}$ instead of $1/N$ as in the case of one-time measurement.
\end{itemize}

\subsection{Race Estimation using BISG}
\label{sec:bisg_overview}
Bayesian Improved Surname and Geocoding (BISG) is a methodology developed by RAND Corporation \cite{BISG} that has been broadly used to calculate estimates of racial and ethnic disparities within datasets where self-provided data is unavailable. BISG relies on public US Census statistics, which are collected, aggregated, and compared with lists of common surnames to establish the probability of a race subgroup given a person's last name and their zip code.

\begin{figure}[htbp]
    \centering
    \includegraphics[width=\columnwidth]{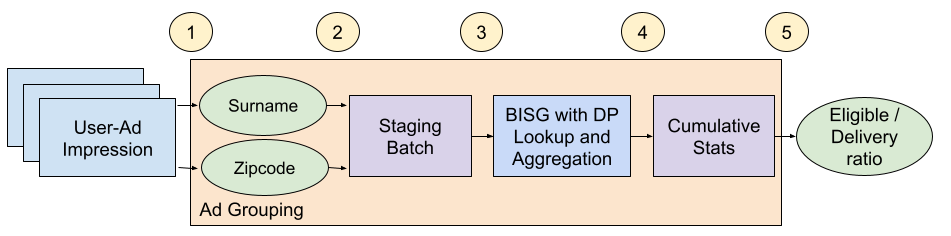}
    \caption{Overview of the BISG + DP pipeline
    }
    \label{fig:dp_overview}
\end{figure}

Our implementation of BISG \cite{BISGMeta} additionally incorporates the aforementioned differential privacy mechanism to help enhance privacy protections. In Figure \ref{fig:dp_overview} and the following steps, we outline an example workflow of how the VRS estimates race or ethnicity. Input data is the impression streaming data. Each data point represents a user-ad impression engagement. Whenever a new data point arrives, we do following:
\begin{enumerate}
    \item Assign it to the group based on the impressed ad id. Each group contains two things: global counter over race buckets and a staging batch.
    \item Add new data's user information (i.e., zipcode and surname) to the staging batch.
    \item When the batch accumulates to a predefined level of aggregation, we send the batch to a dedicated tool implementing BISG with aforementioned privacy features to lookup and aggregate statistics over race buckets.
    \item Then we update the global counter with statistics from staging batch. Staging batch is cleared afterward.
    \item Finally, the eligible/delivery ratio for the ad is computed and returned.

\end{enumerate}

\section{Impression Variance aware Modeling}
\label{sec:Modeling}

As discussed in Section \ref{sec:diff_privacy}, delivery variance can only be measured in an aggregated fashion with DP noise applied. Thus, any feedback for controller decisions is delayed, aggregated, and noisy. Reinforcement Learning (RL) is particularly well suited to solve problems with these constraints, and we elect to use RL as a framework for the VRS model.

When an ad has a chance to be shown as an impression—such as when a user is scrolling through a Meta app and is about to hit a spot where an ad will be displayed—our system gathers ads that would be applicable to show to that user and moves these ads (through ranking stages) to the auction stage where they compete against each other using their "total bid" calculated as:
\begin{equation} \label{total_bid}
	\mbox{\emph{total\char`_bid}} =  { \emph{advertiser\char`_bid}} +  \emph{quality\char`_bid} 
\end{equation}
where \emph{advertiser\char`_bid} represents the monetary value of an impression to the advertiser, and \emph{quality\char`_bid} reflects the user experience associated with an impression.

\begin{figure}[htbp]
    \begin{centering}
    \includegraphics[width=0.95\columnwidth]{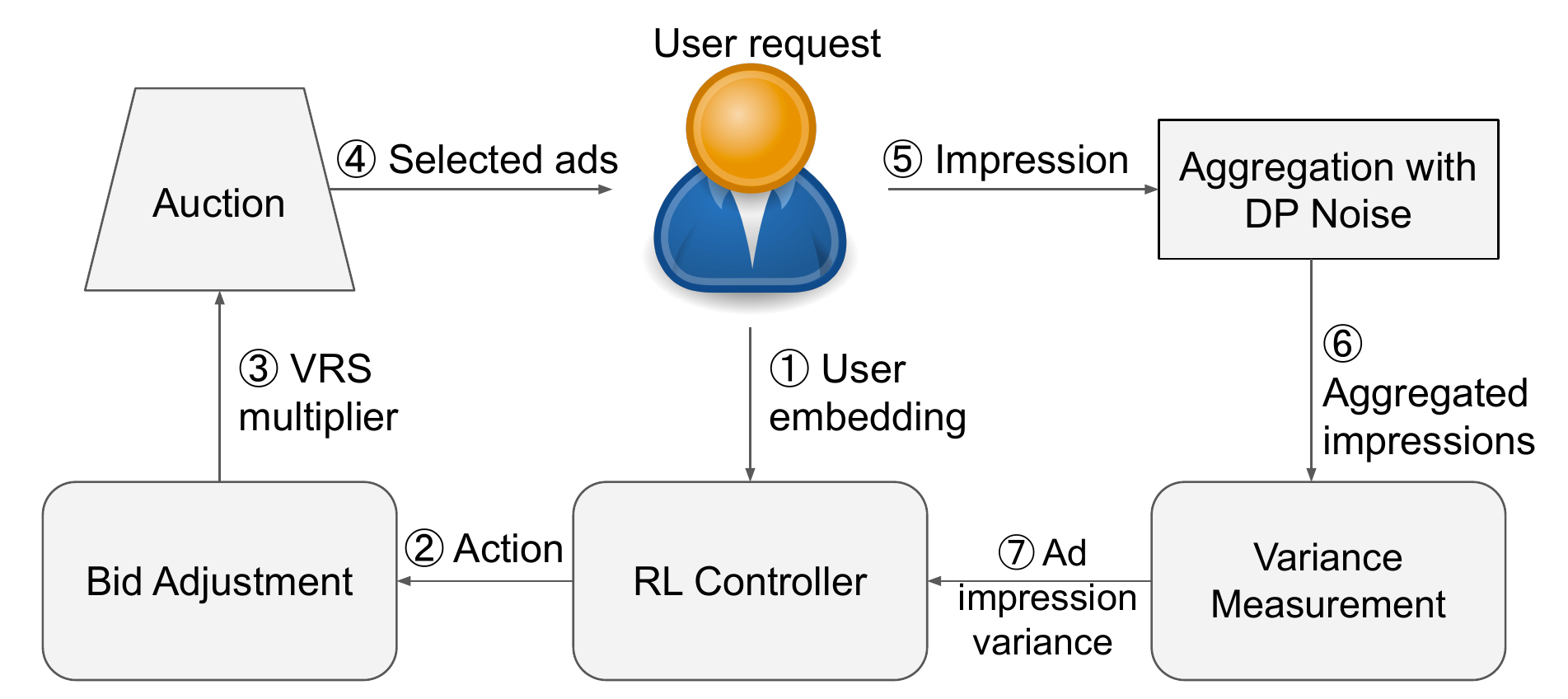}
    \caption{Overview of VRS design}
    \label{fig:rl_controller}
    \end{centering}
\end{figure}

The winning ad will be sent back to the user and shown as an impression.
The VRS relies on an RL controller that has the ability to adjust an ad's total bid in the auction via its \emph{advertiser\char`_bid} component, which will have the effect of changing the likelihood that a given ad will win an auction and be shown to the user (more in Section ~\ref{sec:adjusting_logic}).
The design for the VRS is shown in Figure ~\ref{fig:rl_controller} and summarized in Algorithm ~\ref{alg:rl}. 
By utilizing a user embedding (see Section \ref{sec:user_summarization_model}) and delivery variance information, single-objective RL controllers are trained to control variance for each PC (e.g., gender and estimated race) offline using off-policy data; the policy then remains fixed during evaluation. Single-objective policies are combined via a rule-based hierarchical RL policy, which reduces variance for all PCs simultaneously (more in Section ~\ref{sec:hrl}). Models are evaluated both in a real, production environment and using our simulator (described in Section ~\ref{sec:offline_simulation}). For each user request, when scoring a user-housing ad pair in the auction, a user embedding and a measurement of delivery variance for the ad are provided as input to the RL controller. The RL controller then produces a categorical action which determines how the total bid should be modified by a VRS multiplier to increase or decrease the likelihood of an impression; the exact value of the VRS multiplier is computed by the bid adjustment module discussed in Section \ref{sec:adjusting_logic}. After applying the VRS multiplier, the auction proceeds normally to select the winning ads to send back to the user, and delivery variance measurements are periodically updated using the resulting impression data. This feedback loop is characterized as the Markov Decision Process (MDP) for the RL controller. In the following sections, each component of the VRS is described in greater detail.
\subsection{Reinforcement Learning Controller Design}

This section will discuss the formulation of a single-objective VRS task 
as an MDP, the algorithm and neural network architectures used to solve the MDP, learning via offline, off-policy data. 

\subsubsection{MDP Formulation}
\label{sec:mdp_formulation}

The prevailing problem framework for Reinforcement Learning is the MDP. The MDP is defined by the 5-tuple $\tup{S, A, R, P, \gamma}$. $S$ is the distribution of states, and $A$ is the set of actions. $R: S \times A  \rightarrow \mathbb{R}$ is the reward function, which provides a scalar value describing progress toward an objective. $P : S \times A \rightarrow S \times R$ is the dynamics model, and $\gamma \in [0, 1]$ is the discount factor. The objective agent is to learn a policy $\pi(S_t)$ which maximizes the expected return $E[G_t] = \mathbb{E}\big[\sum_{k=t}^{T} \gamma^{k-t} R_k \big]$ at each step in a series of discrete timesteps $t$. $T \in [0, \inf)$ is the terminal timestep that signals the end of an episode. The remainder of this section will describe the formulation of the problem as an MDP.

\paragraph{States:} Each user-housing-ad pair induces a new timestep. For each auction including a housing ad, the model will utilize state information to select an action. For each user-ad pair, there are two main components to the state vector: (1) A user embedding, learned by the user summarization model (discussed in Section \ref{sec:user_summarization_model}), and (2) a vector measurement of the current delivery variance of the ad  (denoted as $v$). For a model balancing gender delivery variance, the input includes only the sign of the female delivery variance $\sign(v_{f})$ as input, while a race model includes the sign of the delivery variance $\sign(v_{i})$ for all race buckets $i$.
In the training data, the delivery variance is periodically updated after every $k$ impressions to help protect the PC information of individual users.

\paragraph{Actions:} The action space is binary: (1) Adjust-up and (2) No-adjustment. The action is executed by applying a multiplier to the bid price of the advertisement prior to the auction.
Alternatively, the bid adjustment module can implement adjust-down in lieu of no-adjustment. 
The computation of VRS multipliers by the bid adjustment module and the details of the alternative adjust-down action are left to Section \ref{sec:adjusting_logic}, and the rest of this section will assume the action space is $A=\{Adjust{\text -}up, No{\text -}adjustment\}$. 

\paragraph{Rewards:}
\label{sec:rewards}

The reward is defined individually for each PC (e.g. race) as follows:
\begin{equation} 
    R_t =
    \begin{cases} 
      0 & t < T \\
      \sum^{|B|}_{i=1} (count_{i}^{1} - count_{i}^{0} ) \cdot \sign(v_{i}^{0}) \cdot (-1)  & t = T
   \end{cases}
\end{equation}
where $B$ is the set of all buckets in the PC (i.e. Black, Hispanic, White, etc), $count_{i}^{1}$, and  $count_{i}^{0}$ are the number of impressions for bucket $i$ after and before the episode respectively; each have differential privacy noise added, as discussed in Section \ref{sec:diff_privacy}. $v_{i}^{0}$ is the delivery variance of bucket $i$ at the beginning of the episode. If $v_{i}^{0}$ is negative, indicating the bucket is underserved, the term for bucket $i$ will be positive and all impressions to members of bucket $i$ yield a positive reward. In contrast, if $v_{i}^{0}$ is positive, the term for bucket $i$ will be negative. This reward has a simple interpretation: With respect to the most recent delivery variance measurement, applying an adjust-up action to a housing ad for an underserved group is always correct, and applying an adjust-up action to a housing ad for an overserved group is always incorrect. Thus, another interpretation is $R_T = \sum(correct\ adjust{\text -}up\ actions) - \sum(incorrect\ adjust{\text -}up\ actions)$.

\paragraph{Episodes:}

The reward function is computed using the same counters as delivery variance. These counters are updated after $k$ impressions to help protect user privacy. Thus, timesteps are grouped into episodes of length $k$ so that there is at-most one non-zero reward on the final timestep $T$. If $k=1$, it's clear that selecting actions which maximize the one-step reward is optimal and matches our intended behavior: the agent will select adjust-up housing ads to an underserved class to reduce the delivery variance, and will select no-adjustment for housing ads to an overserved class to prevent the delivery variance from worsening. In practice, $k>1$ to help protect user privacy, and it is sufficient to optimize with respect to the next terminal reward to achieve the intended optimal behavior. As a result, dividing data into episodes is justified and does not alter the optimal behavior.

\subsubsection{Data Filtering and Augmentation}
\label{data_preprocessing}

Training data is collected from the simulator (Section ~\ref{sec:offline_simulation}) via a uniform random policy. From a statistical perspective, no-adjustment actions are not expected to result in impressions, and adjust-up actions are expected to result in impressions. The model is trained to maximize the expected rewards. Thus, no-adjustment actions which lead to impressions and adjust-up actions which do not lead to impressions are noise in the dataset and can be safely filtered out without altering the optimal policy.

In addition, we filter out episodes containing at least one no-adjustment actions since those actions are not expected to lead to impressions and hence have no effect on the variance or the reward. 
As a result, all episodes in the training data only consist of adjust-up actions and have the same length $k$ (i.e. the aggregation level). These episodes are placed in a replay buffer, which the algorithm uses to learn in an offline, off-policy fashion. 

Training data collected from a uniform random policy is typically imbalanced with respect to the delivery variance measurement (e.g. there may be more episodes where females are underserved relative to males, or vice versa). We found the model is able to learn more effectively if we mirror each episode by performing the following operations: (1) multiply the input delivery variance features by $-1$, and (2) recompute the reward at timestep $T$ using the mirrored delivery variance. Both the original episodes and mirrored episode are included in the training dataset. For each group, this ensures there are an equal number of samples in which each group is underserved/overserved.

\subsubsection{RL Algorithm}
One notable observation of the MDP we have defined above is as follows: The agent's actions do not influence the transitions into future states. User embeddings depend on a user's activity on the platform, and are independent of the decisions to adjust ads, and delivery variance features are updated only after the final timestep of an episode. Similar to a contextual bandit problem, this implies that greedy behavior with respect to the reward function is sufficient to yield an optimal policy. Thus, traditional RL algorithms such as Q-Learning \cite{watkins1992q} and REINFORCE \cite{williams_simple_1992}, which learn policies that optimize the value function, are unnecessarily complex for this problem. 

However, rewards are aggregated such that the only possible non-zero reward is $R_T$. The hidden, per-step rewards, $R^*_t$, are not observed by the agent. Thus, we propose a method of return decomposition \cite{arjona2019rudder, efroni_reinforcement_2021} to de-aggregate the observed return, $G_t = \sum_{i=t}^T \gamma^{i-t} R_i$, into its hidden per-step components, $R^*_t$. Recall, the alternative interpretation of the reward is $R_T = \sum(correct\ adjust{\text -}up\ actions) - \sum(incorrect\ adjust{\text -}up\ actions)$, and all no-adjustment actions are filtered out of the training data. Thus, each hidden, per-step reward $R^*_t \in \{-1, 1\}$. In this problem formulation, $\gamma=1$. Each $R^*_t$ can be predicted using only the current state and an action. Afterward, the policy selects actions which greedily maximize the learned reward function at evaluation time. 
\sloppy At every update step, the agent receives one trajectory $\tau =(S_0, A_0, R_1, S_1, A_1, R_2,...,S_{T-1}, A_{T-1}, R_T) \sim D$  such that $T=k$, and $D$ is the dataset which was collected via a random policy and preprocessed. 
$R_T$ does have zero-mean DP noise added, and the expected value of the final received reward can be alternatively defined as:

\begin{equation}
    E\big[R_T\big] = E\big[G_0\big] = \sum\limits_{k=1}^T R^*_k
\end{equation}
Recall, the preprocessed trajectories $\tau$ only contain adjust-up actions. As a result, the learned function $f(S | \theta)$ is trained to predict the value of an adjust-up action given state $S$, where $\theta$ are the parameters of a neural network. The optimization problem is as follows:
\begin{equation} \label{eq:optim_problem_rd}
\begin{split}
    J(\tau, \theta) &= \big|\big| \sum\limits_{S \in \tau} f(S |\theta) - R_T \big|\big|_2 \\
\theta^* &= argmin_{\theta}\sum\limits_{\tau \in D}J(\tau, \theta)
\end{split}
\end{equation}
where $\theta^*$ are the optimal parameters of the neural network. Crucially, no-adjustment actions always yield $0$ reward, and the final approximation of the reward function is:
\begin{equation} \label{eq:learned_reward_function}
    R(S, A | \theta) =
    \begin{cases} 
      0 & A = no{\text -}adjustment \\
      f(S | \theta)  & A = adjust{\text -}up
  \end{cases}
\end{equation}
Thus, the optimal policy $\pi^*$ can be defined as:
\begin{equation} \label{eq:policy_def}
\pi^*(S) = argmax_a R(S, a | \theta^*)
\end{equation}
Equation \ref{eq:policy_def} is used to form the final policy given the learned reward function. The full training algorithm is shown in Algorithm \ref{alg:RD}, and the evaluation time algorithm for inferencing the trained policy is shown in Algorithm \ref{alg:rl}. 

\begin{algorithm}[H]
\caption{Reward Function Learning}
\label{alg:RD}

\begin{algorithmic}[1]
    \State \textbf{Input:} Preprocessed Dataset $D$, learning rate $\alpha$
    \State \textbf{Output:} Trained parameters $\theta$
    \State \textbf{Initialize:} Randomly initialize neural network parameters $\theta$
    \While {not converged}
        \State Set $L=0$
        \For {i in range(update frequency)}
            \State Sample a trajectory $\tau \sim D$
            \State Accumulate loss $L = L + J(\tau, \theta)$
        \EndFor
        \State Perform a gradient step $\theta = \theta - \alpha\nabla_{\theta}L$
    \EndWhile
\end{algorithmic}
\end{algorithm}

\subsubsection{Modeling Architecture}

\begin{figure}[htbp]
    \centering
    \includegraphics[width=0.85\columnwidth]{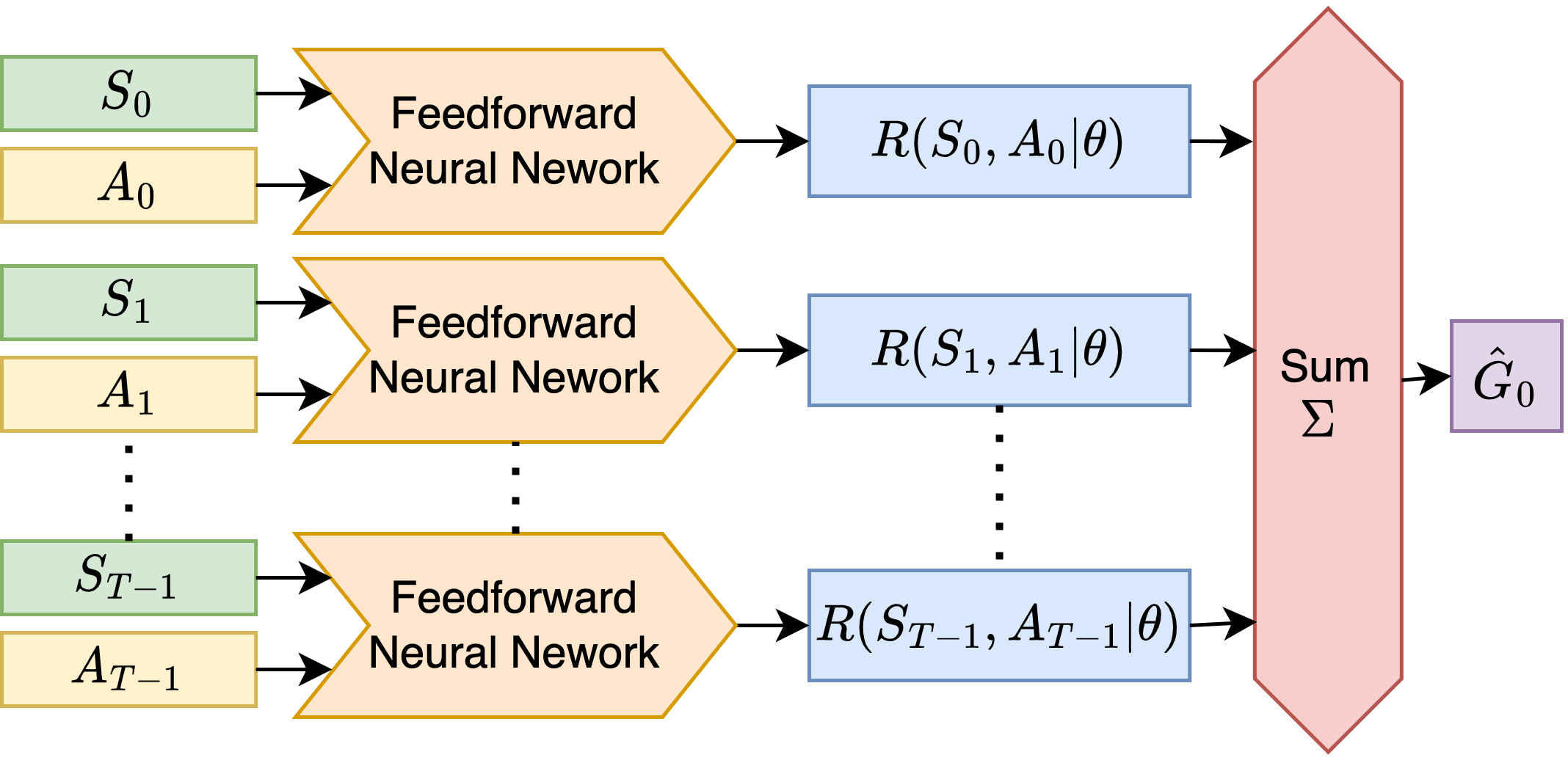}
    \caption{Shared MLP architecture used at train-time. The neural network uses identical parameters to process each S, A pair. The sum of these outputs is an estimate of the return.
    }
    \label{fig:reward_decompose}
\end{figure}

The train-time neural network architecture is shown in Figure \ref{fig:reward_decompose}. Notably, \textbf{the same parameters are used to process each state-action pair}, and one output is produced per pair. Thus, the neural network in Figure \ref{fig:reward_decompose} can be alternatively viewed as one neural network which is used to process $T$ inputs. At train time, the agent computes the estimated return $\hat{G}_0$ by computing the sum of the $T$ corresponding outputs. The estimated return is used to compute the loss in Equation \ref{eq:optim_problem_rd}, where $\hat{G}_0 = \sum\limits_{S \in \tau} f(S |\theta)$ because the training data only contains adjust-up actions. At evaluation time, only the current state $S$ is needed to select the optimal action. The agent computes the value of each possible action, then selects the action which maximizes the estimate of the reward function using the policy defined in Equation \ref{eq:policy_def}.

\subsubsection{Offline Evaluation}
For validating the RL controller's performance during training, we use an offline evaluation metric using the offline simulation data without accessing individual user's PC attributes.
The metric $d$, called "the adjust-up difference", is defined as follows:
\begin{equation} \label{adjustup_metric}
	\mbox{\emph{d}} = \Tilde{x}  - x
\end{equation}
where $\Tilde{x}$ is the number of predicted adjust-up actions by the controller for a given episode, and $x$ is the number of expected adjust-up actions.
A smaller $|d|$ corresponds to better controller performance. 
See Appendix ~\ref{offline_evaluation-appendix} for the derivation of $x$ using simulation data.

\subsection{Multi-objective Modeling}
\label{sec:hrl}

Ultimately, the VRS aims to minimize delivery variance for multiple PCs (e.g. gender and estimated race) simultaneously. To handle multi-objective modeling, a typical approach is to define a separate reward function for each objective, then combine them to engineer a scalar reward function that captures all objectives. Then, a policy would be trained using the multi-objective scalar reward, the resulting policy observed, and then the reward function would be re-engineered if the behavior is not acceptable. This iterative method would be repeated until the behavior is satisfactory to the designer \cite{Hayes_2022}.
\begin{figure}[htbp]
    \centering
    \includegraphics[width=0.85\columnwidth]{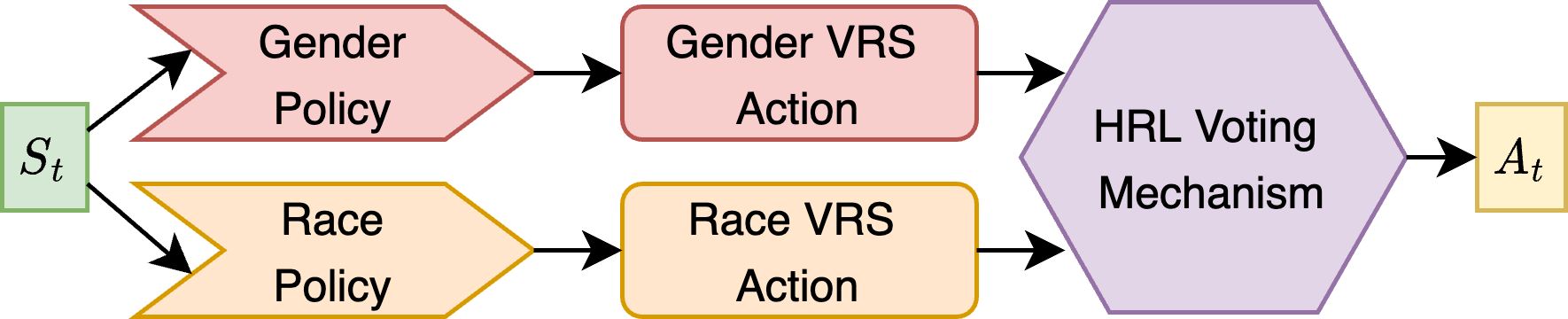}
    \caption{The HRL controller used for inference}
    \label{fig:hrl_controller}
\end{figure}

To solve this multi-objective RL problem, we elect to use a simplified version of Hierarchical Reinforcement Learning (HRL). In this design, a dedicated controller is trained for each PC. At inference time, each single-objective controller is queried for an action, and each single-objective action is used as input to a hand-engineered meta-controller, which outputs the final VRS action. By combining solutions to simpler, single-objective tasks, this hierarchical design is scalable, modular, and is simple to train and tune.

The meta-controller implements an `equal' voting mechanism to determine the action, and is illustrated by Figure \ref{fig:hrl_controller}. The output of each single objective controller is considered a vote for that action. If there is a tie in the vote, the controller is conservative and selects the no-adjustment action. We experimented with other voting schemes in the meta controller, but found that equal voting to be the most effective in practice. See Appendix ~\ref{multi-appendix} for details of other options.

\subsection{User Summarization Model}
\label{sec:user_summarization_model}

\begin{figure}[htbp]
    \begin{centering}
    \includegraphics[width=0.85\columnwidth]{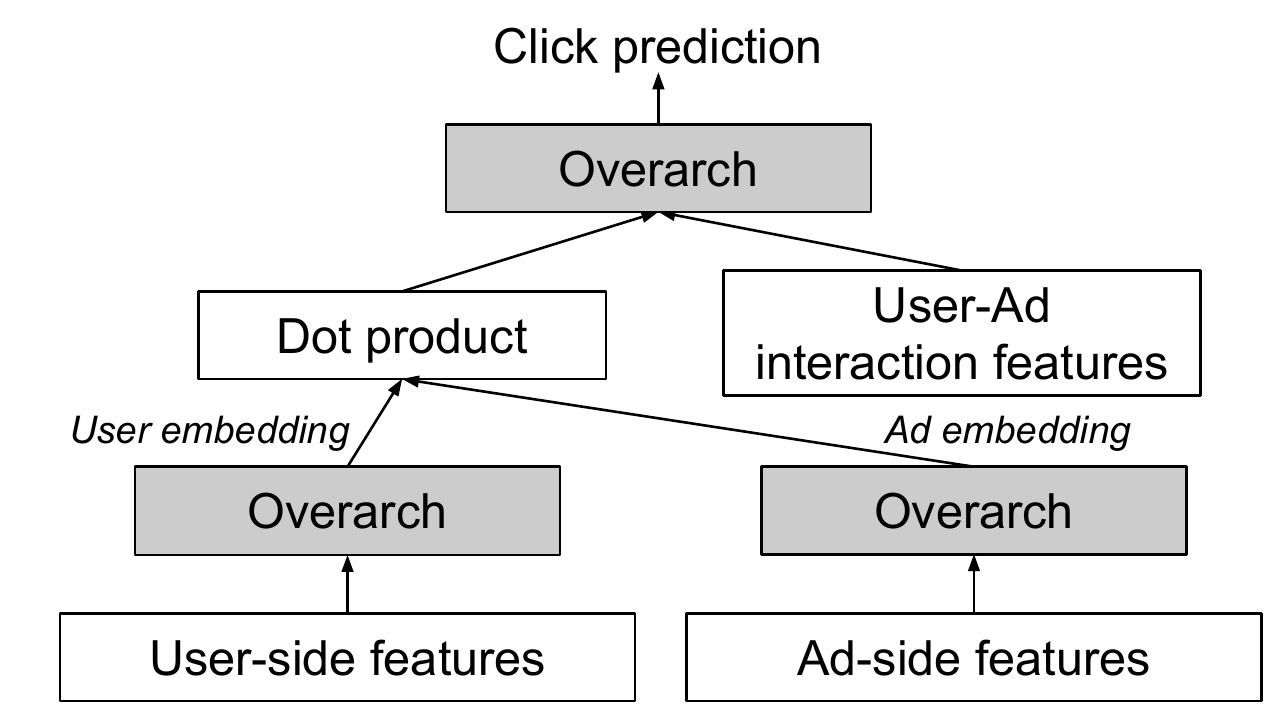}
    \caption{User Summarization Model Architecture}
    \label{fig:user_summarization_model}
    \end{centering}
\end{figure}

The user-side information will be represented as a user embedding for each user using data that excludes their PC information. Specifically, we first train a user summarization model using the clickthrough data on ad impressions as the label and features from user side, ad side, and user-ad interaction as the input. The user summarization model has a two-tower architecture as shown in Figure ~\ref{fig:user_summarization_model}, which includes a user arch (which takes only user-side features as its input), an ad arch (which takes only ad-side features as its input) and the top interaction arch. Then, the user embeddings are extracted from the output of user arch in the trained model at inference time. Such a user summarization model allows us to achieve following goals: (1) Ensure that the generated user embeddings do not directly contain PC attribute-related information by excluding input features that directly contain or are derived from the PC attributes. (2) The user embeddings can be generated for users across Meta's platforms. (3) Lightweight to facilitate fast inference as only the user-arch needs to be evaluated.

\begin{algorithm}[htbp]
\caption{Impression Variance Aware Ads Ranking Stack}
\label{alg:rl}
\begin{algorithmic}[1]
    \For{each \emph{event} in $Events$}
    \State ${user_\char`_request} \gets$ \emph{event} 
    \State $U_{embed} \gets$ $UserEmbedding(user_\char`_request)$
    \State $v_{0} \gets$ Dictionary with key as ad id and value as input variance 
    \State $ads \gets$ list of ad candidates 
    \For{each \emph{ad} in $ads$}
        \State $action \gets$ RLController($v_{0}\{ad\}$, $U_{embed}$) 
        \State $VRS_{multiplier} \gets$ $BidAdjustmentModule($\emph{action}$)$ 
        \State $UpdateTotalBid(\emph{$VRS_{multiplier}$}, \emph{ad})$
    \EndFor
    \State  \textit{\# run simulation to obtain expected impression}
    \State ${impressed_\char`_ads} \gets$ $AdsStack(event)$
    \For{each \emph{ad} in $impressed_\char`_ads$} 
        \State $UpdateVariance(\emph{ad})$
    
    \EndFor
    \EndFor
\end{algorithmic}
\end{algorithm}

We run inference of the trained user summarization model on a daily basis to ensure user embeddings cover new users who just joined the Meta platforms and the freshness of the user embeddings for existing users.
In production, we precompute user embeddings which are served upon each user request to avoid the need to run the user summarization model on the fly.

\subsection{Bid Adjustment Module}
\label{sec:adjusting_logic}

For each housing ad in a given auction, the RL controller outputs one of the two actions: no-adjustment and adjust-up. 
Bid adjustment module then converts these categorical actions to a VRS multiplier that will be applied to \emph{advertiser\char`_bid} of the ad before it enters auction stage and modifies \emph{total\char`_bid} as in Equation \ref{adjusted_total_bid}:
\begin{equation} \label{adjusted_total_bid}
	\mbox{\emph{total\char`_bid}} =  { \emph{advertiser\char`_bid}} * { \emph{vrs\char`_multiplier}} +  \emph{quality\char`_bid} 
\end{equation}
The VRS multiplier is determined for each of the RL controller's action as follows.

\paragraph{No-adjustment} \emph{vrs\char`_multiplier} is set to 1 so that the RL controller has no effect on the total bid of the ad.

\paragraph{Adjust-up}
In this case, one wants to increase the \emph{total\char`_bid} of the ad so that it has a higher chance of winning the auction and becoming an impression, thereby reducing the variance. 
In practice, we apply a fixed VRS multiplier for all housing ads to be adjusted up.
This fixed multiplier is determined by the P40 of the multipliers needed to bring up a housing ad to the top of the auction based on the historical data (so roughly 40\% of the time the adjustment brings the target ad to the top). 

When the action space only includes no-adjustment and adjust-up, the bid adjustment logic is denoted as "Adjust-up Only".
To give the RL controller additional capability to shift the distribution of impressions toward the eligible ratio, we convert the RL controller's no-adjustment action to adjust-down. 
The intuition is that when the RL controller outputs a no-adjustment action, the ad may still become a realized impression with a small probability and therefore worsen the variance. 
Adjust-down helps the ad rank below other candidate ads competing in the auction with higher total bids, and thus further reduces the likelihood it becomes a realized impression. 
\paragraph{Adjust-down} Similar to adjust-up, for adjust-down we use a fixed VRS multiplier which is the P15 of multipliers needed to bring down a housing ad to the bottom of the auction based on the historical data (so roughly 85\% of the time the adjustment brings the target ad to the bottom).

When the action space includes both adjust-up and adjust-down, the bid adjustment logic is denoted "Adjust-up and Down".
We experimented with both Adjust-up Only and Adjust-up and Down logic in the offline simulation and online A/B tests as discussed in Section ~\ref{sec:Results}.
The calculation of multipliers needed to bring a given ad to the top/bottom of the auction is detailed in Appendix~\ref{bid_adjust-appendix}.

\section{Offline Simulation Environment}
\label{sec:offline_simulation}
We developed an offline environment to simulate a simplified version of the ads delivery system in production for data generation, modeling, and offline evaluation. More specifically, this offline simulation environment is designed to re-run the ranking and auction stack given the ads production log data as input. The offline simulation environment, independent of the ads delivery system in production, builds a closed and well-controlled environment to facilitate the modeling development.

\begin{figure}[htbp]
    \begin{centering}
    \includegraphics[width=0.95\columnwidth,trim={1cm 1.25cm 2cm 5.75cm},clip]{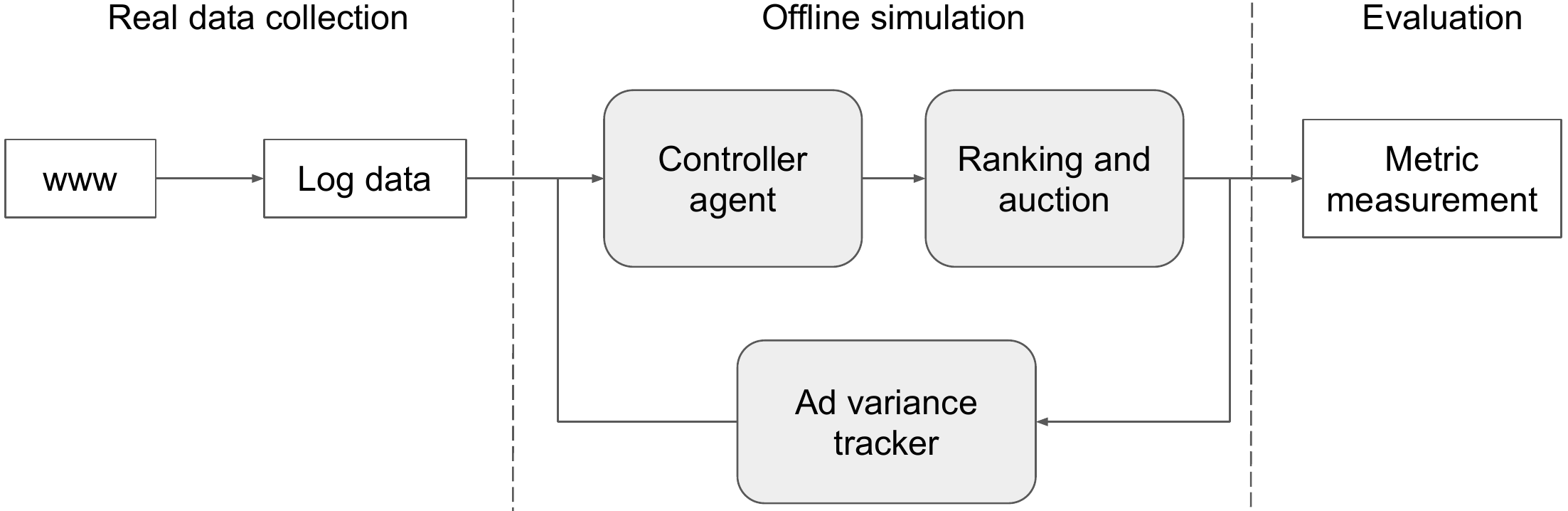}
    \caption{The pipeline of offline simulation environment}
    \label{fig:simulator}
    \end{centering}
\end{figure}

Figure ~\ref{fig:simulator} presents the pipeline of offline simulation environment. We first collect data in the real world from the website. The log data contains the following information:
(1) All candidate ads information. This would ensure we are able to re-rank candidate ads when a VRS multiplier is applied (explained in Section ~\ref{sec:adjusting_logic}).
(2) Users request level information. This includes user-side features that are needed by agent model (e.g., user embedding), as well as other metadata. 
(3) Auction slot information. This is needed to re-run the auction process. Instead of a full detailed information on auction slots, we are mostly interested in the number of available auction slots only in order to simulate the final list of delivered ads. 

The controller agent modifies the total bid of ads candidates in each user request before the auction, according to the user embedding as well as the variance. Then, the ranking and auction stage determines the final list of delivered ads with the modified bid. Ad variance tracker tracks the delivered ads and compute the delivery variance, which is fed back to the controller agent for the next user request. Finally, we collect the simulation results for the metric measurement and performance evaluation.

\section{Results}
\label{sec:Results}

In this section we explain our experimental settings as well as providing results for offline evaluation and online A/B testing.

\subsection{Experimental Settings}

\subsubsection{Metrics}
\paragraph{Coverage}
We define coverage as:
\begin{equation} \label{coverage}
    \mbox{\emph{Coverage}} = \frac{y}{x}
\end{equation}
where $y$ is the total number of ad under a specific variance threshold (e.g., 10\% ) and $x$ is the total count of housing ads. 

\paragraph{Coverage for Shuffle Distance}
To have an aggregated Coverage for PC attributes which are not binary (e.g., race has 4 buckets), we use the shuffle distance to compute the ad variance as defined in Section ~\ref{sssec:elligible}. 

\paragraph{NCAC}
Non-conforming Ad Coverage (NCAC) is the fraction of  ads that are not under a specific variance threshold. This can be obtained by subtracting \emph{Coverage for Shuffle Distance} from 1. 

\paragraph{NCAC Reduction}
We define NCAC reduction as in Equation ~\ref{NCAC_Reduction} as the metric we use for reporting.
\begin{equation}
\label{NCAC_Reduction}
    NCAC\_reduction = (1-\frac{NCAC_{test}}{NCAC_{control}})\cdot100\%
\end{equation}
For reporting, we only consider ads that have at least 300 total impressions.

\subsubsection{Design Choices for PC Subgroups}
We categories estimated race PC attribute into four categories: Black, Hispanic, White, and Other (everything else).
We categories gender PC attribute into male and female.

\subsection{Simulation Results }
The results of our offline evaluation are shown in Table ~\ref{offline_results}. Test 1 and 2 differ in terms of bid adjustment logic. Test 1 is the model with Adjust-up Only, and Test 2 is the model with Adjust-up and Down. Both tests results in significant reduction in NCAC percentage. 
In terms of Test 1 vs Test 2, Test 2 appears to work significantly better than Test 2. 
The intuition is that when the RL controller outputs a `no-adjust' action, 
adjusting down helps the ad lands below other candidate ads competing in the auction with higher total value, and thus reduces the likelihood it becomes a realized impression. More testing is needed to ratify these initial conclusions, but the early results do appear promising for Adjust-up and Down.
\begin{table}[!htb]
\caption{\% in NCAC reduction for 10\%  threshold offline }
\centering
\begin{tabular}{l rr rr}
\toprule


{Test No.} & {Bid adjustment logic}   & {\emph{Gender}}  & {\emph{Race}} \\
\midrule
{Test 1} & {Adjust-up Only} & 84.08 &50.79 & \\ \hline
{Test 2} & {Adjust-up and Down} &92.25 &67.68 & \\
\bottomrule

\end{tabular}
\label{offline_results}
\end{table}
\subsection{Online A/B Testing Results and Deployments in Meta's Ads System}

We performed an A/B testing with a duration of two weeks in Meta's ads system, with two different settings in terms of bid adjustment logic. Similar to offline evaluation, Test 1 is the model with Adjust-up Only, and Test 2 is the model with Adjust-up and Down.
We randomly allocated the same number of ad segments in each treatment group.
The results of  A/B testing  are shown in Table ~\ref{qrt1_results}. The numbers in the table show percentages in NCAC reduction.
As can be seen, our test groups both reduced the NCAC metric with respect to the control group, and the \emph{Adjust-up and Down} showed it might have better variance reduction capability than the \emph{Adjust-up Only}. Similar to offline evaluation, we think this is because the \emph{Adjust-up and Down} gives our system more flexibility to modify the ads ranking to mitigate the potential variance between eligible and delivery ratios. 

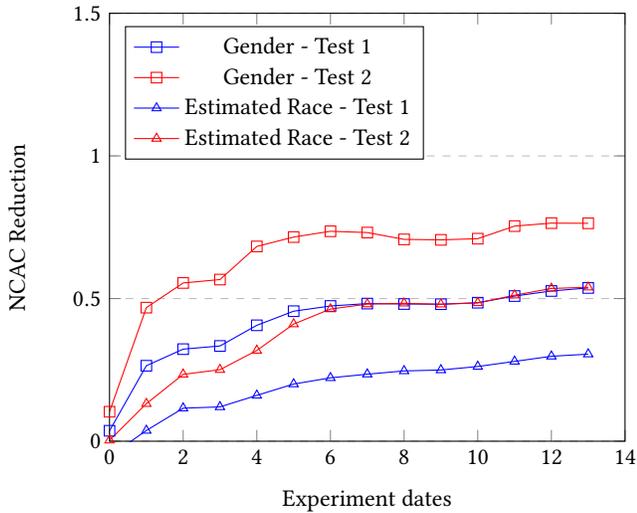
\begin{figure}[!h]
\centering
\begin{tikzpicture}
\begin{axis}[
    title={},
    xlabel={Experiment dates},
    ylabel={NCAC Reduction},
    xmin=0, xmax=14,
    ymin=0, ymax=1.5,
    xtick={},
    ytick={},
    legend pos=north west,
    ymajorgrids=true,
    grid style=dashed,
]

\addplot[
    color=blue,
    mark=square,
    ]
    coordinates {
    (0,0.0366225996931327)(1,0.2645126379604561)(2,0.32271133031306215)(3,0.33383608875598375)(4,0.40625312504734923)(5,0.4556850685222208)(6,0.47393409126256053)(7,0.48236893395933317)(8,0.48085282334081875)(9,0.4802072683092322)(10,0.4854733155111851)(11,0.5087540858056919)(12,0.5265346023488792)(13,0.5371598597279696)
    };
\addlegendentry{Gender - Test 1}

\addplot[
    color=red,
    mark=square,
    ]
    coordinates {
    (0,0.10332429506470808)(1,0.4681696043803762)(2,0.5545908291585411)(3,0.5665481716055144)(4,0.6826534909839646)(5,0.7155179790665689)(6,0.7358319844371357)(7,0.7313396645908814)(8,0.707393554101736)(9,0.7057307271058957)(10,0.7098658305167057)(11,0.7539615315867451)(12,0.764358307455091)(13,0.7637754372692993)
    };
\addlegendentry{Gender - Test 2}

\addplot[
    color=blue,
    mark=triangle,
    ]
    coordinates {
    (0,-0.05313054191274374)(1,0.03752426592024363)(2,0.11565458740705482)(3,0.12021460839417031)(4,0.16062695397073087)(5,0.20032060531569346)(6,0.2220986495628988)(7,0.23492817727867327)(8,0.24632684988734826)(9,0.24960139374737988)(10,0.2618445879215456)(11,0.27973313903634645)(12,0.2976896673179294)(13,0.30518688576780495)
    };
\addlegendentry{Estimated Race - Test 1}

\addplot[
    color=red,
    mark=triangle,
    ]
    coordinates {
    (0,0.003367003367003339)(1,0.13144030261269998)(2,0.23433303831602526)(3,0.25066411691595714)(4,0.3180236324745252)(5,0.4105497349030955)(6,0.46324520947952824)(7,0.48058941403571204)(8,0.48344949142535215)(9,0.4798739228615014)(10,0.4847738540939974)(11,0.5116638256108269)(12,0.5353140668388469)(13,0.539564808663717)
    };
\addlegendentry{Estimated Race - Test 2}
    
\end{axis}
\end{tikzpicture}
\caption{NCAC reduction vs. online A/B experiment dates for Test 1 (Adjust-up Only) and Test 2 (Adjust-up and Down)}
\label{fig:qrt1_result}
\end{figure}

\begin{table}[hbt!]
\caption{\% in NCAC reduction for 10\%  threshold online }
\label{tab:delay}
\centering
\begin{tabular}{l rr rr}
\toprule


{Test No.} & {Bid adjustment logic}  & {\emph{Gender}}  & {\emph{Race}} \\
\midrule
{Test 1} & {Adjust-up Only} & 53.77 &30.45 & \\ \hline
{Test 2} & {Adjust-up and Down} &76.36 &53.95 & \\
\bottomrule

\end{tabular}
\label{qrt1_results}
\end{table}
Notably, online performance in the online A/B tests is degraded relative to the offline simulation tests. This is the result of overhead added by the extensive infrastructure required to support online serving, which is not present in offline simulation. Additionally, the simulator does not perfectly emulate online auction dynamics. As the VRS infrastructure is improved, we expect this gap to shrink. We leave further work on reducing the offline-to-online gap as future work.
We also show how NCAC reduction increases over the experiment dates in Figure ~\ref{fig:qrt1_result}. In general, we observe a sharp increase at the first few days. Then the increase will slow down and eventually converge to some plateau. We find that it usually takes two weeks for the result to stabilize.

\section{Related Work}
\label{sec:Related Work}
To the best of our knowledge, our proposed framework is the first large-scale privacy-preserving framework deployed in production to improve fairness for multiple PC attributes.  
There is a large and growing body of literature on developing mathematical fairness criteria according to different societal and ethical notions of fairness, as well as techniques for building machine learning models that fulfil those fairness criteria. The fairness enforced machine learning algorithms in the literature mainly fall into three categories. In the following, we review each. 
For a comprehensive survey of fairness in machine learning we refer the reader to \citep{mehrabi2021survey, friedler2019comparative, hort2022bia, li2022fairness, wang2022survey}.

\textbf{1) Pre-processing}. Pre-processing methods typically aim to remove bias in data. 
The studies by~\citet{zemel2013learning,calmon2017optimized} map the data to a fair representation in a latent space satisfying the defined fairness or independence of the PC attribute to remove potential bias in the data, and then use the mapped data in the subsequent machine learning. ~\citet{lahoti2020fairness} proposed Adversarially Reweighted Learning to achieve Rawlsian Max-Min fairness in which a learner is trained to optimize performance on a classification task while the adversary adjust the wights of computationally-identifiable regions in the input space with high training loss.

\textbf{2) In-processing}.  In-processing methods aim at encoding fairness as part of the objective function (i.e., constraining the training process). 
The works by \citet{ge2021, Ge2022TowardPE} studied the problem of fairness on the item exposure in recommendation system by proposing a fairness-constrained reinforcement algorithm which can achieve the Pareto frontier of fairness and utility, and thereby facilitate decision makers to control fairness utility trade-off.
\citet{Wu2021} studied user and provider fairness in news recommendation with adversarial learning and propose to learn a biased-aware user/provider embedding and a bias-free user/provider embedding simultaneously and encourage them to be orthogonal to each other. 
\citet{wang2020robust} explored the practical problem of enforcing group-based fairness for binary classification using noisy PC attribute without a privacy guarantee. 
The works by~\citet{zafar2017eo,zafar2017dp} constrain the classifier on demographic parity and equalized opportunity, respectively, during the training. In a follow up work \citet{zafar2017parity} introduced preference-based notion of fairness. Their methods train a linear classifier for each group by solving a coupled empirical risk minimization problem that enforces preferences with a convex surrogate loss function.
The work by ~\citet{pmlr-v124-chen20b} introduces the equalized distribution, a stronger fairness condition, on the demographic parity and equalized opportunity concepts to alleviate the sensitivity of the trained ``fair'' model to the decision threshold tuning. In these methods, the tradeoff is usually required between accuracy and fairness constraints. 

\textbf{3) Post-processing}. Post-processing methods tend to modify the presentations of the results. Our framework VRS categorizes as a post-processing method. 
\citet{Sahin2019} explored several re-ranking rules to provide fair ranking scores and the desired proportions over gender attribute. 
\citet{do2021two} define the notion of fairness in increasing the utility of the worse-off individuals and proposed an algorithm based on maximizing concave welfare functions using Frank-Wolfe algorithm. 
The work by~\citet{hardt2016equality} searches proper thresholds on the model output scores over different groups to obtain the fairness compliant classification. The study by~\citet{pleiss2017fairness} calibrates on the prediction score such that the probability of positive label is equal to the prediction score. 

Additionally, our proposed return decomposition algorithm is not described precisely in the literature, to the best of our knowledge. However, both \cite{efroni_reinforcement_2021} and \cite{ren_learning_2022} describe similar algorithms which attempt to regress the per-step reward function using aggregated feedback. 

\section{Conclusion and Future Work}
\label{sec:Discussion}
Motivated by improving equitable access to livelihood opportunities to all Meta users, we proposed the VRS (a post-processing method) to help ensure the audience that ends up seeing a housing ad more closely reflects the eligible targeted audience for the ad and hence equitable outcome. We verified the effectiveness of the VRS in reducing potential bias, in the form of ad impression variance, via extensive simulation evaluations and A/B Testing and achieved significant improvements with regard to our fairness metric. We deployed the VRS for pursuing fairness in Meta’s delivery of US housing ads for gender and estimated race PC attributes. VRS is the first large-scaled deployed framework for pursuing fairness in the personalized ad domain for multiple PC attributes. We are currently working towards deploying the VRS for credit and employment ads. As we expand VRS to an additional PC attribute and new verticals, learning the HRL policy from data remains a promising direction for future work.

VRS represents several years of progress in consultation with advocates, researchers, policymakers and other stakeholders. Much of this work is unprecedented in the advertising industry and represents a significant technological advancement for how machine learning is responsibly used to deliver personalized ads. We additionally hope that our work could be of interest for researchers and practitioners working on similar domains.


Beyond our advertising system we continue to pursue work to embed both civil rights considerations and responsible AI into our product development process, some of which was shared in Meta’s civil rights audit progress report released in late 2021. We know that our ongoing progress — both in ads fairness and broader civil rights initiatives — will be determined not just by our commitment to this work, but by concrete changes we make in our products. We look forward to not only building solutions, but participating in and supporting the critical, industry-wide conversations that lie ahead.

\begin{acks}
We would like to thank Andrew Howard and Isabella Leone for their ongoing work to implement targeting policies to help prevent discriminatory audiences and support in building the VRS, as well as the Ads Fairness, Responsible AI, and Integrity teams. Additionally, we would like to thank Damien Wint from our Legal team for his support. Finally we would like to thank Renfei Chen, Kellie Lu and Samuel Drews for their contributions during the early stage of the project.



\end{acks}

\clearpage

\bibliographystyle{ACM-Reference-Format}
\balance
\bibliography{ camera_ready_revised_kdd}

\clearpage
\appendix

\section{Modeling}

\subsection{Multi-objective Modeling}
\label{multi-appendix}

For voting weight, we additionally tried two other options. 
In the first option, we weigh each action from each controller according to the absolute values of current variance of each bucket within each PC, as shown in equation \ref{shuffle_voting}:

\begin{equation} \label{shuffle_voting}
	\mbox{Vote(PCX)} = \frac{ \sum^{buckets}_{i} \lvert variance(PCX_{i})\rvert }{\sum^{PCs}_{X} \sum^{buckets}_{i} \lvert variance(PCX_{i})\rvert} \times a_{x}
\end{equation}

In the second option, we weigh each action from each controller according to the maximum absolute value of current variance of each bucket within each PC as shown in equation \ref{max_voting}:

\begin{equation} \label{max_voting}
	\mbox{Vote(PCX)} = \frac{ \max_{i \in {buckets}} \lvert variance(PCX_{i})\rvert }{\sum^{PCs}_{X} \max_{i \in {buckets}} \lvert variance(PCX_{i})\rvert} \times a_{x}
\end{equation}
Our experimental results showed equal voting gives best performance in practice.

\subsection{Bid Adjustment Module}
\label{bid_adjust-appendix}
To bring the \emph{total\char`_bid} of a given ad to the top of the auction, the \emph{vrs\char`_multiplier} can be determined by the Equation \ref{adjusting_up}:
\begin{equation} \label{adjusting_up}
	\emph{vrs\char`_multiplier} = \frac{\mbox{\emph{max total\char`_bid}} -  \emph{quality\char`_bid}} { \emph{advertiser\char`_bid}}
\end{equation}
where \emph{max total\char`_bid} is the maximum \emph{total\char`_bid} among all ads in the same auction (housing and non-housing), and \emph{quality\char`_bid} and \emph{advertiser\char`_bid} are of the given ad.

To bring the ad to the bottom of the auction one simply needs to replace \emph{max total\char`_bid} in Equation \ref{adjusting_up} with \emph{min total\char`_bid}:
\begin{equation} \label{adjusting_down}
	\emph{VRS\char`_multiplier} = \frac{\mbox{\emph{min total\char`_bid}} -  \emph{quality\char`_bid}} { \emph{advertiser\char`_bid}}
\end{equation}
where \emph{min total\char`_bid} is the minimum \emph{total\char`_bid} among all ads in the same auction.

\subsection{Offline Evaluation Metric}
\label{offline_evaluation-appendix}

To compute the expected number of adjust-up actions in an episode ($x$), recall that after data filtering all $k$ actions in a training episode are adjust-up actions (see Section ~\ref{data_preprocessing}).
As a result, the expected number of adjust-up actions equals to the correct number of adjust-up actions in a training episode.
Also recall that the episode reward is defined as the difference in correct number of adjust-up actions (i.e., $x$) minus incorrect number of adjust-up actions in an episode (see Section ~\ref{sec:mdp_formulation}), we derive the episode reward $R$ as
\begin{equation} \label{reward}
	\mbox{\emph{R}} = x - (k - x) = 2x-k
\end{equation}

Therefore, we can compute $x$ using the episode reward and the episode length in the training data as follows:

\begin{equation} \label{expected_adjustup}
	\mbox{\emph{x}} = \frac{R + k}{2}
\end{equation}

\end{document}